# Masked Conditional Neural Networks for Environmental Sound Classification

Fady Medhat    David Chesmore    John Robinson

Department of Electronic Engineering
University of York, York
United Kingdom
`{fady.medhat,david.chesmore,john.robinson}@york.ac.uk`

**Abstract.** The ConditionaL Neural Network (CLNN) exploits the nature of the temporal sequencing of the sound signal represented in a spectrogram, and its variant the Masked ConditionaL Neural Network (MCLNN)[1] induces the network to learn in frequency bands by embedding a filterbank-like sparseness over the network's links using a binary mask. Additionally, the masking automates the exploration of different feature combinations concurrently analogous to handcrafting the optimum combination of features for a recognition task. We have evaluated the MCLNN performance using the Urbansound8k dataset of environmental sounds. Additionally, we present a collection of manually recorded sounds for rail and road traffic, YorNoise, to investigate the confusion rates among machine generated sounds possessing low-frequency components. MCLNN has achieved competitive results without augmentation and using 12% of the trainable parameters utilized by an equivalent model based on state-of-the-art Convolutional Neural Networks on the Urbansound8k. We extended the Urbansound8k dataset with YorNoise, where experiments have shown that common tonal properties affect the classification performance.

## 1    Introduction

Automatic feature extraction for signals either image or sound is gaining a wide interest from the research community aiming to eliminate the efforts invested in handcrafting the optimum features for a recognition task. Inspired by the generative Deep Belief Nets (DBN) introduced by Hinton et al. [1], the deep architectural structure got adapted to

---

[1] Code: https://github.com/fadymedhat/MCLNN

This work is funded by the European Union's Seventh Framework Programme for research, technological development and demonstration under grant agreement no. 608014 (CAPACITIE).

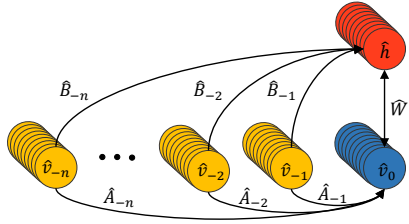 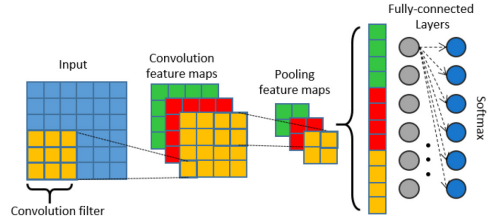

**Fig. 1.** The Conditional RBM structure     **Fig. 2.** Convolutional Neural Network

other models such as the Convolutional Neural Networks (CNN) [2]. These deep architectures are used to extract an abstract representation of the features that can be further classified using a conventional classifier, e.g. SVM [3]. An attempt to use a neural network based architecture over the raw sound signal has been considered by Dieleman et al. in [4]. Their work endeavored to bypass the need for spectrograms as an intermediate signal representation, but their findings showed that spectrograms prevail. An attempt to use stacked Restricted Boltzmann Machines [5] forming a DBN for the task of music genre classification was in the work of Hamel et al. [6], where the extracted features were classified using an SVM. Despite the successful attempts in using the DBN as a feature extractor for sound, it treats each spectrogram frame as an isolated entity. This ignores the temporal relation across the frames.

To capture the successional relationship in sequential data, extending from an RBM structure, Taylor et al. proposed the Conditional RBM (CRBM) [7]. Taylor used the CRBM for modeling the human motion through tracking the joints position of the human body as features across time. The CRBM was applied for sound signals such as drum pattern analysis in [8]. The CRBM as shown in Fig. 1 accounts for the interframes relation of a temporal signal by including conditional links from the previous visible nodes $(\hat{v}_{-n}, \ldots, \hat{v}_{-2}, \hat{v}_{-1})$ to both the hidden layer $\hat{h}$ and the current visible input $\hat{v}_0$. The Interpolating CRBM (ICRBM) [9], a variant of the CRBM, enhanced the speech phoneme recognition accuracy compared to the CRBM by considering future frames in addition to the past ones.

The Convolutional Neural Network (CNN) [2], shown in Fig. 2, is another very successful model that achieved breakthrough results in image recognition [10]. The CNN operates using two consecutive layers of convolution and pooling operations. In the convolutional layer, the input is scanned with an array of filters to generate a number of feature maps matching the number of filters. Then a pooling layer is used to compress the resolution of the convolutional layer's output either through mean or max pooling. Several of these two layers can be stacked on top of each other in a deep architectural structure, where the output of the last convolutional or pooling layer can be fed into a dense layer for the final classification decision. The CNN depends on weight sharing, which does not preserve the spatial locality of the learned feature. The locality of the features detected in a spectrogram is crucial for distinguishing between sounds, which induced attempts [11],[12] to tailor the CNN filters for sound recognition to capture both the temporal and spectral properties in addition to tackling the translation invariance property of the CNN.

Long Short-Term Memory (LSTM) [13] is a Recurrent Neural Network (RNN) architecture that allows considering a past sequence of temporal frames using internal memory. Choi et al. [14] exploited the use of a hybrid model in the Convolutional RNN, where they used a CNN to extract the local features and an LSTM to capture the long-term dependencies of a temporal signal for music.

The Convolutional DBN (ConvDBN) [15] was an extension to the generative models based on the RBM. The ConvDBN investigated the use of the unsupervised training of an RBM for images by adapting the convolutional behavior of a CNN and through introducing a probabilistic pooling layer. ConvDBN was adopted in [16] for music and speech tasks.

We have highlighted the most relevant architectures to this work. Models were developed for other applications, primarily image recognition, and then adapted to the nature of the sound signal. This may not optimally harness the multidimensional representation of an audio signal as a spectrogram. For example, the DBN ignores the inter-frame relations of a temporal signal, where it treats the frames as isolated entities in a Bag-of-Frames classification. Also, models based on the convolution operation like the CNN depends on weight sharing, which permits CNNs to scale well for images of large dimensions without having a dedicated weight for each pixel in the input. Weight sharing makes the CNN translation invariant, which does not preserve the spatial locality of the learned features. The ConditionaL Neural Networks (CLNN) [17] and its variant the Masked ConditionaL Neural Network (MCLNN) [17] are developed from the ground up exploiting the nature of the sound signal. The CLNN considers the inter-frames relation in a temporal signal and the MCLNN embeds a filterbank-like behavior that enables individual bands and suppresses others through an enforced systematic sparseness. Additionally, the mask in the MCLNN automates the exploration of different feature combinations concurrently, which is usually a handcrafted operation of finding the optimum feature combinations through exhaustive trials. Meanwhile, the MCLNN preserves the spatial locality of the learned features. In this work, we extend the evaluation of the MCLNN in [18] to the Urbansound8k dataset. Additionally, we investigate the confusion across machine generated sounds possessing common low tonal components through YorNoise, a dataset we are presenting in this work focusing on rail and road traffic sounds.

## 2  Conditional Neural Networks

Sound frames can be classified one frame at a time, but higher accuracy is achieved by exploiting the relationship between the frames across time, as in the CRBM discussed earlier. A CRBM is a generative model possessing directed links between the previous frames and both the current visible and hidden layers as shown in Fig. 1. These links hold the conditional relation of observing a particular pattern of neurons' activations at either of the hidden or visible layer conditioned on the previous $n$ visible vectors. These directed links convert an RBM into the Conditional RBM.

A CLNN captures the temporal nature of a spectrogram and allows end-to-end discriminative training by extending from the CRBM using the past $n$ visible-to-hidden links in addition to the future ones as proposed in the ICRBM. This allows the network to learn from the temporal data and acts as the main skeleton for the MCLNN.

For notation purposes, matrices are represented with uppercase symbols with the hat operator $\widehat{W}$ and vectors with lowercase symbols $\hat{x}$. $\widehat{W}_u$ is the matrix at index $u$ in the tensor $\widehat{W}$. $W_{i,j,u}$ is the element at location $[i, j]$ of a matrix $\widehat{W}_u$, similarly $x_i$ is the $i^{th}$ element of the vector $\hat{x}$. The dot operator ( · ) refers to vector-matrix multiplication. Element-wise multiplication between vectors or matrices of the same sizes uses ( ∘ ). The absence of any operators or the use ( × ) refers to normal elements multiplication ($l \times e$ or $x_i\, W_{i,j}$).

The CLNN hidden layer is formed of vector-shaped neurons and accepts an input of size $[l, d]$, where $l$ is the length of the feature vector and $d$ is the number of frames in a window following (1)

$$d = 2n + 1 \quad , n \geq 1 \tag{1}$$

where the order $n$ controls the number of frames in a single temporal direction, $2n$ is for the frames on either side of the window's central frame, and 1 is for the middle frame itself. Accordingly, at any temporal instance, the CLNN hidden layer activations are conditioned on the window's central frame and the $2n$ neighboring frames. A single input vector within the window is fully connected with the hidden layer having $e$-neurons. The dense connections between each vector and the hidden layer are captured in a dedicated weight matrix $\widehat{W}_u$, where $u$ is the index of the weight matrix residing in the weight tensor. The index $u$ ranges within $[-n, n]$ matching the number of frames $d$. The activation of hidden neuron follows (2)

$$y_{j,t} = f\left(b_j + \sum_{u=-n}^{n} \sum_{i=1}^{l} x_{i,u+t}\, W_{i,j,u}\right) \tag{2}$$

where $y_{j,t}$ is the activation of the $j^{th}$ neuron, $f$ is the activation function, $b_j$ is the bias at the $j^{th}$ neuron, $x_{i,u+t}$ is the $i^{th}$ feature in the feature vector at index $u+t$, where $u$ is the index of the vector in the window, and $W_{i,j,u}$ is the weight between the $i^{th}$ feature of the feature vector at index $u$ in the window and the $j^{th}$ neuron of the hidden layer, where each frame at index $u$ has a corresponding weight matrix of the same index in the weight tensor. The $t$ index in the equation refers to the index of the frame in a sequence of frames, which we will refer to as the segment (discussed later in detail). Accordingly, the frame at index $t$ within the segment is the window's central frame $\hat{x}_{u+t}$ at $u = 0$. The vector formulation of (2) is given in (3)

$$\hat{y}_t = f\left(\hat{b} + \sum_{u=-n}^{n} \hat{x}_{u+t} \cdot \widehat{W}_u\right) \tag{3}$$

where $\hat{y}_t$ is the predicted activations of the window's middle frame $\hat{x}_t$ ($u = 0$) conditioned on $2n$ off-central frames within $[-n+t, n+t]$, $f$ is the transfer function, $\hat{b}$ is the bias vector at the hidden layer, $\hat{x}_{u+t}$ is the feature vector (having length $l$) within the window at index $u$, where $\hat{x}_t$ is the frame at index $t$ of the segment and also the window's central frame at $u = 0$, and $\widehat{W}_u$ is the weight matrix (having a size [feature vector $l$, hidden layer length $e$]) at index $u$ corresponding to the vector at index $u+t$. The generated $d$ vectors from the vector-matrix multiplication are summed per dimension to generate a single vector to apply the transfer function on. The generated vector is a representative frame for the input window of frames. The conditional distribution of the frames can be captured using a sigmoid function at the hidden layer or through the final output softmax through $p(\hat{y}_t| \hat{x}_{-n+t}, ..., \hat{x}_{-1+t}, \hat{x}_t, \hat{x}_{1+t}, ..., \hat{x}_{n+t}) = \sigma(...)$, where $\sigma$ is the transfer function used.

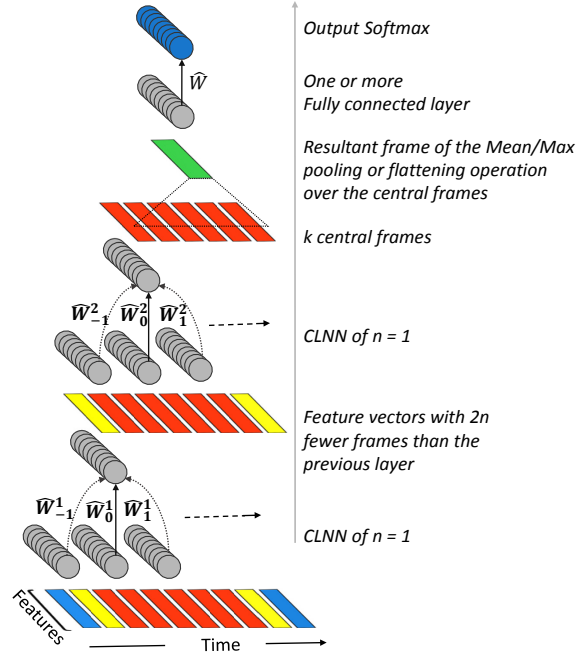

**Fig. 3.** A two layer CLNN model with *n*=1

According to (3), the output of a CLNN step over a window of *d* frames is a single vector $\hat{y}$. This highlights the consumption of the frames in a CLNN layer, where the output is 2*n* frames fewer than the input. To account for such reduction of frames in a deep CLNN architecture, a segment of frames is fed to the deep architecture following (4)

$$q = (2n)m + k \quad , n, m \text{ and } k \geq 1 \qquad (4)$$

where *q* is the width of the segment, *n* is the order (multiplied by 2 to account for past and future frames) controlling the width of the window at a single CLNN layer, *m* is for the number of layers and *k* is for the extra frames. *k* specifies the number of frames that should remain beyond the CLNN layers to be flattened to a single vector or globally pooled [19] across the temporal dimension. This behaves as an aggregation over a texture window that was studied in [20] for music classification. Finally, the generated vector is introduced to a densely connected neural network for the final classification.

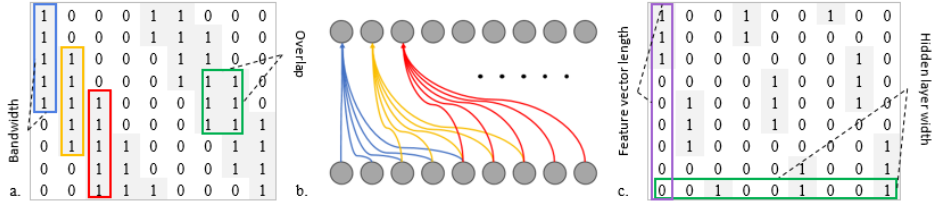

**Fig. 4.** Examples of the Mask patterns. a) A bandwidth of 5 with an overlap of 3, b) The allowed connections matching the mask in a. across the neurons of two layers, c) A bandwidth of 3 and an overlap of -1

Fig. 3 shows a deep CLNN structure composed of two layers, *m*=2, and having an order *n*=1. In this setting, each frame is conditioned on one previous and one succeeding frame. Each CLNN layer is composed of a weight tensor $\widehat{W}^b$, where *b* is layer index (*b*=1, 2, ..., *m*). The center weight matrix $\widehat{W}_0^b$ processes the main frame at time *u+t* at *u* = 0 and an additional *2n* weight matrices, where at *n*=1 there are $\widehat{W}_{-1}^b$ to handle one previous frame and $\widehat{W}_1^b$ to handle for the future one. The figure also depicts the remaining *k* frames to be flattened or pooled across before the densely-connected layers.

## 3   Masked Conditional Neural Networks

The Masked ConditionaL Neural Networks (MCLNN) [17] use the same structure and behavior as the CLNN and additionally enforces a systematic sparseness over the network's connections through a masking operation.

Spectrograms have been used widely as an intermediate signal representation. They allow an in-depth analysis of the signal structure and the frequency components forming the signal. They describe the change in the energy assigned to each frequency bin as the signal progresses through time. However, they have some shortcomings for sound recognition. The energy of the different frequencies of a sound signal is affected by environmental acoustic factors during the signal propagation. These factors may cause the energy to be shifted from one frequency bin to a nearby frequency bin for the same sound signal, resulting in a different feature vector to a recognition system. Filterbanks are used in this regard to subdivide the frequency spectrum into bands, which provide a frequency shift-invariant representation. A Mel-Scaled filterbank is a principle operating block used in time-frequency representations of sound in Mel-Frequency Cepstral Coefficients (MFCC) or Mel-Scaled spectrograms.

The MCLNN embeds a filterbank-like behavior through the utilization of a binary mask. The mask is a binary matrix matching the size of a weight matrix as shown in Fig. 4.a., where the positions of the 1's are arranged based on two parameters: the Bandwidth *bw* and the Overlap *ov*. The bandwidth controls the number of frequency bins to be considered together, and the overlap controls the superposition distance between bands. Fig. 4.a depicts a bandwidth of 5 across the rows and an overlap between the bands (across the columns) equal to 3. Fig. 4.b shows the active connections matching the mask pattern defined in Fig. 4.a. The overlap can be assigned negative values, which refer to the non-overlapping distance between the successive bands as shown in Fig. 4.c. The mask design is based on a linear spacing that follows (5)

$$lx = a + (g-1)(l + (bw - ov)) \qquad (5)$$

where the linear index $lx$ of a position of a binary value 1 is given by the bandwidth $bw$ and the overlap $ov$. The values of $a$ are within the interval $[0, bw\text{-}1]$ and the values of $g$ are in the interval $[1, \lceil (l \times e)/(l + (bw - ov)) \rceil ]$.

The mask suppresses the weights in the 0's locations by elementwise multiplication, enforcing a systematic sparseness over the network's connections between the input to any layer and the scope of the interest of each hidden node as formulated in (6).

$$\hat{Z}_u = \widehat{W}_u \circ \widehat{M} \qquad (6)$$

where $\widehat{W}_u$ is the original weight matrix and $\widehat{M}$ is the mask pattern. $\hat{Z}_u$ is the masked weight matrix to substitute $\widehat{W}_u$ in (3).

Another important role of the mask is the process of automating the exploration of a range of feature combinations concurrently similar to the manual hand-crafting of different feature combinations. This is implemented in the mask through the presence of several shifted versions of the filterbank-like binary pattern. For example, in Fig. 4.c. (columns map to the hidden layer neurons), the 1st neuron will learn about the first three features in the input feature vector (ignoring the temporal dimension for simplicity) matching the positions of the three 1's present in the mask. The 4th neuron will learn about the first two features, and the 7th neuron will learn about the first feature only. This behavior allows different neurons to observe a focused region of features in the feature vector while preserving the spatial locality of the learned features.

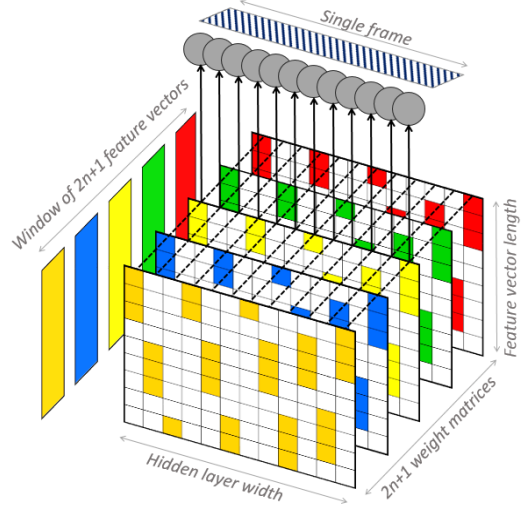

**Fig. 5.** A single step of MCLNN

Fig. 5 shows a single step of the MCLNN, where $2n+1$ frames have a matching number of $2n+1$ matrices. The highlighted regions represent the active connections following the mask design. The output of a single MCLNN step processing the $2n+1$ input frames is a single representative frame.

## 4   Experiments

We performed the experiments using the Urbansound8k [21] and the YorNoise ( presented in this work) datasets of environmental sounds. The Urbansound8k dataset is composed of 8732 audio sample of 10 categories of urban sounds: air conditioner, car horns, children playing, dog bark, drilling, engine idling, gunshot, jackhammers, siren and street music. The maximum duration of each file is 4 seconds. The dataset is released into 10-folds. We used the same arrangement of folds to unify the benchmarking and eliminate the data split influence on the reported accuracies. We will defer the YorNoise discussion to its relevant experiments later in this section.

All files were re-sampled to a monaural channel of 22050 Hz sampling rate and 16-bit word depth wav format. The files were transformed to a logarithmic Mel-scaled spectrogram of 60 bins using an FFT window of 1024 and a 50% overlap and with their delta (first derivative of the spectrogram along the temporal dimension). The spectrogram and the delta were concatenated to generate a spectrogram of 120 bin. Following the time-frequency transformation, segments were extracted following (4). All training files are standardized feature-wise using z-scoring, and the standardization parameters (mean and standard deviation) of the training folds were applied on both the validation and test folds.

For the MCLNN model, Parametric Rectifier Linear Units (PReLU) [22] were used as activation functions. The MCLNN layers are followed by a global mean pooling layer [19], but for a temporal signal such as sound, it is a feature-wise single-dimensional pooling across the $k$ extra frames. Finally, the fully-connected layers are used before the softmax output. We used two densely connected layers of 100 neurons each. The network was trained to minimize the categorical cross-entropy using ADAM [23], and Dropout [24] was used to prevent overfitting. The final decision of a clip's category was based on a probability voting across the frames. We used Keras[1] and Theano[2] on a GPU for the model's implementation, and the signal transformation was carried out using LibROSA[3] and FFmpeg[4]. Table 1 lists the hyper-parameters used for the MCLNN model we adopted for the Urbansound8k dataset.

Table 1. MCLNN model parameters for Urbansound8k

| Layer | Type | Nodes | Mask Bandwidth | Mask Overlap | Order $n$ |
|---|---|---|---|---|---|
| 1 | MCLNN | 300 | 20 | -5 | 15 |
| 2 | MCLNN | 200 | 5 | 3 | 15 |

The model has two MCLNN layers having an order $n = 15$ (Future work will consider different order across the layers). We noticed through several experiments that having a wider bandwidth in the first layer compared to the second layer provides better performance. The role of the mask in the first layer is to process sub-bands within the spectrogram and the narrower bandwidth in the second layer's mask focuses on distinctive features over a small region of bins. The same applies for the overlap, where more sparseness is required in the first layer, through the negative overlap, to eliminate the noisy effect of smearing bins in proximity to each other. The single dimensional global mean pooling was used to pool across $k = 5$ extra frames and finally the fully connected layers for classification.

---

[1] https://keras.io

[2] http://deeplearning.net/software/theano

[3] https://librosa.github.io/

[4] http://ffmpeg.org/

**Table 2.** Accuracies reported on the Urbansound8k

| Classifiers and features | Acc.% |
|---|---|
| Random Forest + Spherical K-Means + PCA + Mel-Spectrogram [25] | 73.7 |
| **MCLNN + Mel-Spectrogram (This Work)** | **73.3** |
| Piczak-CNN + Mel-Spectrogram [26] | 73.1 |
| S&B-CNN + Mel-Spectrogram [27] | 73.0 |
| RBF-SVM + MFCC [21][1] | 68.0 |

Table 2 lists the mean accuracy achieved by the MCLNN across the 10-fold cross validation, in addition to other methods applied on the Urbansound8k dataset. The MCLNN achieved an accuracy of 73.3%, which is the highest neural based accuracy for the Urbansound8k dataset. The highest non-neural accuracy is 73.7%, reported in the work of Salamon [25], used the random forest as a classifier applied over a dictionary established using Spherical K-Means [28]. As an intermediate representation, they used a Mel-scaled spectrogram dimensionally reduced by PCA.

To be able to benchmark the MCLNN compared to other proposed CNN models having a similar depth, we adopted the same spectrogram transformation used for the CNN proposed by Piczak in [26] (60 bin mel-scaled spectrogram with its Delta). Piczak used a separate channel for each of the spectrogram and the Delta to train a CNN. To fit this to the MCLNN, we concatenated both transformations column-wise, resulting in a frame size of 120 features. Piczak experimented with two segment sizes extracted from the spectrogram to train a CNN, where a short segment is composed of 41 frames, and a long segment comprises 101 frames. The highest accuracy reported in [26] was 73.1% using a long segment variant to train the Piczak-CNN. The CNN proposed by Piczak in [26] is composed of two convolution layers of 80 filters each, two pooling layers followed by two fully-connected layers of 5000 neurons each and finally the output Softmax. The number of trainable weights in Piczak-CNN exceed 25 million parameters, on the other hand, the two MCLNN layers we used for this work required approximately 3 million parameters trained over segments of 65 frames. The deeper CNN architecture proposed by Salamon in [27] used fewer parameters. Accordingly, we will consider deeper MCLNN architectures for future work. MCLNN achieved comparable results to state-of-the-art attempts using 12% of the network parameters of a CNN having a similar depth. The work of Salamon in [27] proposed the use of an augmentation stage by applying different modification, e.g. pitch shifting, time stretching, to the input signal to increase the dataset and consequently enhance the model generalization. Piczak [26] reported the absence of a performance gain when applying augmentation on the Urbansound8k dataset. We did not consider augmentation as it is not relevant to benchmarking the MCLNN performance against other proposed models.

Figure 6 shows the confusion matrix for the Urbansound8k using the MCLNN. The highest confusion is between the Air Conditioner, Drilling, Engine Idling and Jackhammer. This is related to the presence of common low tones across the four classes. Similar confusion was reported in [27].

|   | AC | CH | CP | DB | Dr | EI | GS | Ja | Si | SM |
|---|---|---|---|---|---|---|---|---|---|---|
| AC | **447** | 4 | 50 | 69 | 62 | 169 | 8 | 119 | 14 | 58 |
| CH | 6 | **343** | 10 | 3 | 24 | 7 | 0 | 5 | 3 | 28 |
| CP | 13 | 1 | **845** | 34 | 25 | 17 | 3 | 0 | 12 | 50 |
| DB | 22 | 12 | 88 | **817** | 9 | 7 | 6 | 1 | 13 | 25 |
| Dr | 28 | 9 | 25 | 19 | **752** | 21 | 4 | 97 | 22 | 23 |
| EI | 68 | 8 | 25 | 4 | 43 | **668** | 3 | 148 | 15 | 18 |
| GS | 0 | 0 | 1 | 15 | 2 | 2 | **351** | 0 | 1 | 2 |
| Ja | 88 | 8 | 1 | 0 | 153 | 93 | 0 | **602** | 40 | 15 |
| Si | 9 | 0 | 72 | 24 | 12 | 10 | 0 | 5 | **764** | 33 |
| SM | 24 | 5 | 125 | 15 | 8 | 2 | 0 | 6 | 17 | **798** |

Classes: Air Conditioner(AC), Car Horns(CH), Children Playing(CP), Dog Bark(DB), Drilling(Dr), Engine Idling(EI), Gun Shot(GS), Jackhammers(Ja), Siren(Si) and Street Music(SM)

**Fig. 6.** Confusion matrix for the Urbansound8k using the MCLNN

We wanted to further analyze the effect of the common low tonal components across the machine generated sounds. For this analysis, we introduce YorNoise[1], a dataset focusing on urban generated sounds especially road vehicles and trains. We collected the sound samples from different locations within the city of York in the United Kingdom. The sound files were recorded using a professional recorder fixed at an altitude of one meter above the ground. The captured mono files were recorded at a 44100 Hz sampling rate, with 5 minutes for each recording on average. Each 5-minute file is split into multiple samples of 4 seconds each matching the setting of the Urbansound8k. We examined every 4 seconds file and cleared disrupted samples and silent ones. The files for both categories are distributed across 10-folds while making sure that the 4 seconds samples belonging to the same 5-minute file are residing in the same fold. The total number of files is 907 of road traffic sounds and 620 for rail.

We applied the same preprocessing used for the Urbansound8k to YorNoise. We appended the 10-folds of YorNoise to the Urbansound8k, which generated a dataset of 12 categories totaling to 10259 sound files (Urbansound8k categories in addition to YorNoise: Rail and Traffic). We applied the same model used for the Urbansound8k to the 12 sound categories. MCLNN achieved a mean accuracy of 75.13% for a 10-folds cross-validation with the confusion shown in Fig. 7. Despite the high recognition accuracy of 95.6% and 97.5% for rail and traffic, respectively, it is clear that YorNoise's categories are either confused among themselves (due to the similarity between the train engine and road vehicles) or with the low tonal classes of the

---
[1] https://github.com/fadymedhat/YorNoise

Urbansound8k (the Air Conditioner, Drilling, Engine Idling and Jackhammer). This further validates the noticeable confusion rates among the low tonal classes of the Urbansoun8k.

|    | AC  | CH  | CP  | DB  | Dr  | EI  | GS  | Ja  | Si  | SM  | Ra  | Tr  |
|----|-----|-----|-----|-----|-----|-----|-----|-----|-----|-----|-----|-----|
| AC | **376** | 3   | 45  | 83  | 145 | 138 | 1   | 63  | 7   | 55  | 64  | 20  |
| CH | 10  | **341** | 5   | 3   | 21  | 11  | 0   | 7   | 1   | 27  | 2   | 1   |
| CP | 17  | 3   | **821** | 35  | 27  | 12  | 2   | 1   | 12  | 53  | 11  | 6   |
| DB | 24  | 12  | 97  | **793** | 12  | 7   | 5   | 1   | 13  | 28  | 3   | 5   |
| Dr | 36  | 3   | 17  | 20  | **757** | 22  | 7   | 73  | 27  | 18  | 5   | 15  |
| EI | 75  | 3   | 22  | 11  | 27  | **654** | 2   | 112 | 9   | 18  | 42  | 25  |
| GS | 0   | 0   | 1   | 16  | 1   | 6   | **346** | 0   | 1   | 1   | 2   | 0   |
| Ja | 100 | 0   | 1   | 0   | 121 | 110 | 1   | **601** | 3   | 16  | 44  | 3   |
| Si | 12  | 0   | 55  | 31  | 3   | 18  | 0   | 5   | **748** | 43  | 1   | 13  |
| SM | 14  | 10  | 115 | 15  | 7   | 8   | 0   | 9   | 19  | **790** | 12  | 1   |
| Ra | 4   | 0   | 0   | 0   | 0   | 5   | 0   | 2   | 1   | 0   | **593** | 15  |
| Tr | 3   | 0   | 0   | 0   | 1   | 0   | 0   | 0   | 1   | 0   | 18  | **884** |

Classes: Air Conditioner(AC), Car Horns(CH), Children Playing(CP), Dog Bark(DB), Drilling(Dr), Engine Idling(EI), Gun Shot(GS), Jackhammers(Ja), Siren(Si), Street Music(SM), Rail (Ra) and Traffic (Tr)

**Fig. 7.** Confusion matrix for the Urbansound8k and YorNoise using the MCLNN

## 5    Conclusions and Future Work

The ConditionaL Neural Network (CLNN) and its extension the Masked ConditionaL Neural Network (MCLNN) are designed for multi-dimensional temporal signals representations. The CLNN considers the inter-frames relation across a temporal signal, and the MCLNN embeds a filterbank-like behavior within the network through an enforced systematic sparseness over the network's links allowing the network to learn in bands rather than bins. Additionally, the presence of several shifted versions of the filterbank-like pattern automates handcrafting the optimum combination of features. MCLNN has shown competitive results compared to state-of-the-art Convolutional Neural Networks on the Urbansound8k environmental sounds dataset. We investigated the confusion across sounds of low tonal component mainly machine generated sounds, through the YorNoise dataset, we introduce in this work, focusing on rail and road traffic. Future work will consider further optimization to the MCLNN architecture and the hyperparameters used. We will also consider multi-channel temporal signals other than sound.

# References

1. G.E. Hinton, R.R. Salakhutdinov: Reducing the Dimensionality of Data with Neural Networks. Science 313, 504-507 (2006)
2. Y. LeCun, L. Bottou, Y. Bengio, P. Haffner: Gradient-based learning applied to document recognition. Proceedings of the IEEE 86, 2278-2324 (1998)
3. B.E. Boser, I.M. Guyon, V.N. Vapnik: A Training Algorithm for Optimal Margin Classifiers. In: Proceedings of the fifth annual workshop on Computational Learning Theory, COLT. (1992)
4. S. Dieleman, B. Schrauwen: End-To-End Learning For Music Audio. In: International Conference on Acoustics, Speech and Signal Processing, ICASSP. (2014)
5. S.E. Fahlman, G.E. Hinton, T.J. Sejnowski: Massively Parallel Architectures for AI: NETL, Thistle, and Boltzmann Machines. In: National Conference on Artificial Intelligence, AAAI. (1983)
6. P. Hamel, D. Eck: Learning Features From Music Audio With Deep Belief Networks. In: International Society for Music Information Retrieval Conference, ISMIR. (2010)
7. G.W. Taylor, G.E. Hinton, S. Roweis: Modeling Human Motion Using Binary Latent Variables. In: Advances in Neural Information Processing Systems, NIPS, pp. 1345-1352. (2006)
8. E. Battenberg, D. Wessel: Analyzing Drum Patterns Using Conditional Deep Belief Networks. In: International Society for Music Information Retrieval, ISMIR. (2012)
9. A.-R. Mohamed, G. Hinton: Phone Recognition Using Restricted Boltzmann Machines In: IEEE International Conference on Acoustics Speech and Signal Processing, ICASSP. (2010)
10. A. Krizhevsky, I. Sutskever, G.E. Hinton: ImageNet Classification with Deep Convolutional Neural Networks. In: Neural Information Processing Systems, NIPS. (2012)
11. O. Abdel-Hamid, A.-R. Mohamed, H. Jiang, L. Deng, G. Penn, D. Yu: Convolutional Neural Networks for Speech Recognition. IEEE/ACM Transactions on Audio, Speech and Language Processing 22, 1533-1545 (2014)
12. J. Pons, T. Lidy, X. Serra: Experimenting with Musically Motivated Convolutional Neural Networks. In: International Workshop on Content-based Multimedia Indexing, CBMI. (2016)
13. S. Hochreiter, J. Schmidhuber: Long short-term memory. Neural Comput 9, 1735-1780 (1997)
14. K. Choi, G. Fazekas, M. Sandler, K. Cho: Convolutional Recurrent Neural Networks for Music Classification. In: arXiv preprint arXiv:1609.04243. (2016)
15. H. Lee, R. Grosse, R. Ranganath, A.Y. Ng: Convolutional deep belief networks for scalable unsupervised learning of hierarchical representations. In: Proceedings of the 26th Annual International Conference on Machine Learning, ICML, pp. 1-8. (2009)
16. H. Lee, Y. Largman, P. Pham, A.Y. Ng: Unsupervised Feature Learning for Audio Classification using Convolutional Deep Belief Networks. In: Neural Information Processing Systems (NIPS). (2009)
17. F. Medhat, D. Chesmore, J. Robinson: Masked Conditional Neural Networks for Audio Classification. In: International Conference on Artificial Neural Networks (ICANN). (2017)
18. F. Medhat, D. Chesmore, J. Robinson: Masked Conditional Neural Networks for Automatic Sound Events Recognition. In: IEEE International Conference on Data Science and Advanced Analytics (DSAA). (2017)
19. M. Lin, Q. Chen, S. Yan: Network In Network. In: International Conference on Learning Representations, ICLR. (2014)
20. J. Bergstra, N. Casagrande, D. Erhan, D. Eck, B. Kégl: Aggregate Features And AdaBoost For Music Classification. Machine Learning 65, 473-484 (2006)

21. J. Salamon, C. Jacoby, J.P. Bello: A Dataset and Taxonomy for Urban Sound Research. In: Proceedings of the 22nd ACM International Conference on Multimedia, pp. 1041-1044. (2014)
22. K. He, X. Zhang, S. Ren, J. Sun: Delving Deep into Rectifiers: Surpassing Human-Level Performance on ImageNet Classification. In: IEEE International Conference on Computer Vision, ICCV. (2015)
23. D. Kingma, J. Ba: ADAM: A Method For Stochastic Optimization. In: International Conference for Learning Representations, ICLR. (2015)
24. N. Srivastava, G. Hinton, A. Krizhevsky, I. Sutskever, R. Salakhutdinov: Dropout: A Simple Way to Prevent Neural Networks from Overfitting. Journal of Machine Learning Research, JMLR 15, 1929-1958 (2014)
25. J. Salamon, J.P. Bello: Unsupervised Feature Learning for Urban Sound Classification. In: IEEE International Conference on Acoustics, Speech, and Signal Processing (ICASSP). (2015)
26. K.J. Piczak: Environmental Sound Classification with Convolutional Neural Networks. In: IEEE international workshop on Machine Learning for Signal Processing (MLSP). (2015)
27. J. Salamon, J.P. Bello: Deep Convolutional Neural Networks and Data Augmentation for Environmental Sound Classification. IEEE Signal Processing Letters (2016)
28. I.S. Dhillon, D.S. Modha: Concept Decompositions for Large Sparse Text Data Using Clustering. Machine Learning, vol. 42, pp. 143-175 (2001)